%% file: Hybrid_WACV.tex
\documentclass[10pt,twocolumn,letterpaper]{article}

\usepackage{wacv}
\usepackage{times}
\usepackage{epsfig}
\usepackage{graphicx}
\usepackage{amsmath}
\usepackage{amssymb}
\usepackage{dsfont}
\usepackage{tablefootnote}
\usepackage{subfig}
\usepackage{boldline} 

\newcommand{\red}[1]{{\color{red}{#1}}}

\usepackage{color,soul}
\usepackage{balance}
\usepackage{multirow}
\usepackage{multicol}

\renewcommand{\eqref}[1]{{{Eq. \ref{#1}}}}

\newif\ifsubmission

\ifsubmission
    \newcommand{\moc}[1]{}
    \newcommand{\mot}[1]{}
    \newcommand{\xiaonote}[1]{}
    \newcommand{\hcxnote}[1]{}
\else
    \newcommand{\moc}[1]{[{\noindent\color{red}#1}]}
    \newcommand{\mot}[1]{{\noindent\color{magenta}#1}}
    \newcommand{\xiaonote}[1]{{\noindent\color{blue}#1}}
    \newcommand{\hcxnote}[1]{{\noindent\color{blue}#1}}
\fi

\def\wacvPaperID{466} 

\wacvfinalcopy 
\ifwacvfinal
\fi

\ifwacvfinal
\usepackage[breaklinks=true,bookmarks=false]{hyperref}
\else
\usepackage[pagebackref=true,breaklinks=true,colorlinks,bookmarks=false]{hyperref}
\fi
\usepackage[capitalise]{cleveref}
\ifwacvfinal
\else
\fi

\begin{document}
\title{MUSCLE: Strengthening Semi-Supervised Learning Via Concurrent Unsupervised Learning Using Mutual Information Maximization
}

\author{Hanchen Xie$^1$, Mohamed E. Hussein$^{1,2}$, Aram Galstyan$^1$, Wael Abd-Almageed$^1$ \\
$^1$USC Information Sciences Institute\\
$^2$Alexandria University, Alexandria, Egypt\\
{\tt\small \{hanchenx,  mehussein, galstyan, wamageed\}@isi.edu}
}

\maketitle

\input{sections/00_abstract}
\input{sections/01_introduction}

\input{sections/02_related_work}
\input{sections/03_prelims}

\input{sections/04_method}

\input{sections/05_eval}
\input{sections/06_conclusion}

\section{Acknowledgements}
This material is based on research sponsored by Air Force Research Laboratory (AFRL)
under agreement number FA8750-19-1-1000. The U.S. Government is authorized to reproduce and distribute reprints for Government purposes notwithstanding any copyright notation therein. The views and conclusions contained herein are those of the authors and should not be interpreted as necessarily representing the official policies or endorsements, either expressed or implied, of Air Force Laboratory, DARPA or the U.S. Government.
{\small
\bibliographystyle{ieee_fullname}
\bibliography{egbib}
}

\end{document}

%% file: sections/00_abstract.tex
\begin{abstract}
Deep neural networks are powerful, massively parameterized machine learning models that have been shown to perform well in supervised learning tasks. However, very large amounts of labeled data are usually needed to train deep neural networks. Several semi-supervised learning approaches have been proposed to train neural networks using smaller amounts of labeled data with a large amount of unlabeled data. The performance of these semi-supervised methods significantly degrades as the size of labeled data decreases. We introduce Mutual-information-based Unsupervised \& Semi-supervised Concurrent LEarning (\mbox{MUSCLE}), a hybrid learning approach that uses mutual information to combine both unsupervised and semi-supervised learning. MUSCLE can be used as a stand-alone training scheme for neural networks, and can also be incorporated into other learning approaches. We show that the proposed hybrid model outperforms state of the art on several standard benchmarks, including CIFAR-10, CIFAR-100, and Mini-Imagenet. Furthermore, the performance gain consistently increases with the reduction in the amount of labeled data, as well as in the presence of bias. We also show that MUSCLE has the potential to boost the classification performance when used in the fine-tuning phase for a model pre-trained only on unlabeled data.
\end{abstract}

%% file: sections/01_introduction.tex
\section{Introduction}
\label{sec:intro}
Over the past decade, Deep Neural Networks (DNN) have been extensively employed and studied in various machine learning domains \cite{bert, speech}. DNNs have become  the standard backbone for solving virtually all computer vision problems, such as image classification \cite{alexnet, vgg, resnet}, object detection \cite{yolo, rcnn, fasterRCNN}, image segmentation \cite{seg1, seg2, seg3}, and human motion prediction \cite{motion_tayper, motion_xiao}. However, due to their massive capacities, DNNs are infamous for requiring large amounts of labeled data.

In the traditional supervised learning paradigm, large amounts of labeled data are essential for training well-performing models. To address this limitation, \textit{few-shot adaptation}~\cite{maml, miniimagenet, prototype}, has been studied. In this approach, using a handful of labeled data samples, a model that has been trained on a similar domain can be adapted to a new domain without compromising the performance on its original domain. While few-shot adaptation is an effective approach, the similarity between the original domain and the novel domains, and the generality of the source model -- which requires a large amount of training data in the original domain -- are crucial for its success.  The commonly used evaluation protocols for few-shot adaptation use class-based splits of a single dataset (i.e. same domain) to create the original and novel domains~\cite{fc100, miniimagenet, Tiered-Imagenet, Fairfs}. Semi-supervised learning (SSL) has been introduced \cite{TemporalEnsembling, meanteacher, fixmatch} to leverage the massive amounts of available unlabeled data, instead of solely relying on labeled data.

\begin{figure}[t]
\begin{center}
   \includegraphics[width=1.0\linewidth]{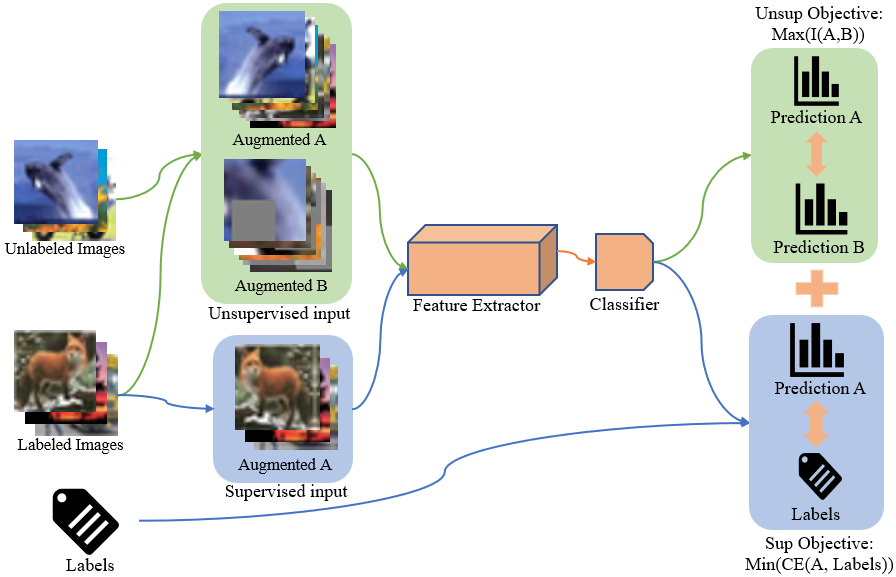}
\end{center}
   \caption{The Training Structure of MUSCLE}
\label{fig:long}
\label{fig:onecol}
\end{figure}

Without loss of generality, SSL can generally be categorized into methods that use  \textit{consistency loss}~\cite{TemporalEnsembling, meanteacher, sajjadi2016regularization, miyato2017virtual} and methods that use \textit{pseudo labeling}~\cite{elezi2018transductive, PSLee, Shi_2018_ECCV, lowshot}. Although these two approaches are orthogonal, combining them was shown to achieve better performance than each one of them individually~\cite{LabelProp2019, fixmatch}. 
In both approaches, knowledge about the task is learned from  labeled samples and transferred to unlabeled samples. In the case of pseudo labeling, unlabeled samples are explicitly labeled (using hard or soft labels), and assigned labels are used to provide supervisory signal in subsequent learning iterations. In the case of consistency loss, the label assignment or feature representation of unlabeled data is forced to be consistent across different models trained simultaneously or different variations of each sample. Despite their success, both approaches share two main weaknesses: 
(1) If the amount of labeled data significantly drops, i.e.  knowledge about the task becomes too limited, they can either fall into degenerate solutions or fail to assign labels with high confidence to much of the unlabeled portion of the training data, and 
(2) Bias in the labeled portion can have a significantly negative impact on the models' performances. This is indeed a problem in all machine learning techniques. However, it can be immensely magnified when few labeled samples are available.


On the other hand, unsupervised learning (USL) techniques extract knowledge from the data without using any labels. Therefore, it is reasonable to assume that combining unsupervised and  semi-supervised learning brings together the best of the two approaches.
Some existing studies attempted at combining USL with Supervised Learning (SL). In such studies, USL can be used as a \textit{pre-training} step, where the model trained via USL is either fine-tuned \cite{DeepInfoMap, iic} or frozen during the SL training~\cite{iic}. Furthermore, in~\cite{learnConnections}, which trains the network layer by layer, USL is used to train the layers and SL is used to learn the connection weights between layers. In these scenarios, the performance of the final SL model may only have a limited increment or even a drop compared to the same model trained directly on the labeled data. This is due to the possible contradiction between the USL's and the SL's training objectives, and hence, the lack of synergy between their two models during training.

In this paper, we show that concurrently using an USL objective along with a SSL model  achieves better performance. We introduce Mutual-information-based Unsupervised \& Semi-supervised Concurrent LEarning (\mbox{MUSCLE}). \mbox{MUSCLE} naturally involves an USL objective, which  maximizes the mutual information between the predictions of variants of the same sample, from the very beginning of the training process. On one hand, when the amount of labeled data is limited, MUSCLE uses the USL objective to gain knowledge about the task. On the other hand, when there is ample labeled data, MUSCLE relies on its SSL objective, while the USL objective can work as a regularization term.
MUSCLE can be used as a stand-alone SSL method, and can also be added to an existing SSL approach. We show that combining MUSCLE with three of the leading SSL approaches~\cite{fixmatch, meanteacher, LabelProp2019} consistently improves the performance on all evaluated benchmarks. Furthermore, the performance gain achieved by MUSCLE consistently increases as the amount of labeled data decreases. We also show that MUSCLE makes the SSL model less sensitive to data bias. Moreover, we show that MUSCLE can be useful in fine-tuning a model pre-trained only on unlabeled data. We provide a thorough discussion about the reasons for such combination to work, and ablation studies on different design parameters to better explain the inner working of the model.


%% file: sections/02_related_work.tex
\section{Related Work}
As mentioned in Section \ref{sec:intro}, there are two main approaches for semi-supervised learning based on either \textit{consistency loss} or \textit{pseudo labeling}.

\textbf{Consistency Loss} has been well studied and included in many SSL techniques ~\cite{TemporalEnsembling, meanteacher, sajjadi2016regularization, miyato2017virtual}.
The basic form of the consistency loss can be expressed as: 
\begin{equation} \label{consistency_loss}
    L_{Consist} = \frac{1}{N}\sum_{i=1}^{N}l_c(f_{\theta}(x_i), f_{\theta'}(x_i'))
\end{equation}
where $f_{\theta}$ is a classification function, $x_i'$ is a variant of the input sample $x_i$, and $l_c$ is a measure of divergence between $f_{\theta}(x_i)$ and $f_{\theta'}(x_i')$, such as $L_1$ or $L_2$ distance~\cite{sajjadi2016regularization, meanteacher}, Jensen-Shannon divergence~\cite{DCT}, and KL-divergence. The source of the variation between $x_i'$ and $x_i$ can be data augmentation~\cite{fixmatch, UDA}, different network parameters~\cite{TemporalEnsembling, meanteacher}, or the randomness inside the network, e.g. dropout \cite{dropout} or noise~\cite{miyato2017virtual}. The basic idea of the consistency loss is that, in the absence of a ground truth label for an input sample $x_i$, the model ensures that variations of the same sample are consistently predicted. However,  consistency loss must be accompanied with a supervised learning loss. Otherwise, we will end up with the  trivial solution in which $f_{\theta}(x_i)$ and $f_{\theta'}(x_i')$ take one value for all classes.

\textbf{Pseudo Labeling}, on the other hand, explicitly assigns labels to unlabeled data, such that the pseudo-labeled data can be used to train regular supervised learning methods, e.g. using cross entropy loss. In~\cite{facebookDeep}, for example, $K$ nearest neighbors ($K$-NN) was used to assign labels to unlabeled samples based on their proximity to labeled samples.
Then, SL, using cross entropy loss, was applied repeatedly to update the model and refine the labels until convergence. Such \textit{hard labeling} approach 
provides a performance gain compared to using only supervised learning on the labeled data as it makes use of the unlabeled data. However, the gain can be limited due to the poor accuracy of the hard-assigned pseudo labels. \textit{Soft labeling}~\cite{LabelProp2019} assigns confidence weights to the pseudo labels to  reduce the negative impact of incorrect pseudo labels.

Completely unsupervised methods have also been suggested for classification tasks without using any labeled data~\cite{IMSAT, iic, DeepInfoMap, learnConnections}. Dundar et al.~\cite{learnConnections} proposed using the k-means clustering algorithm for learning the layers and the connections between layers. Mutual information has also been used in~\cite{DeepInfoMap, iic} for unsupervised classification.

%% file: sections/03_prelims.tex
\section{Preliminaries}
\label{sec:prelims}
In this section, we explain the three leading SSL techniques, which are used as baseline for our proposed method. Beyond the basic idea of the consistency loss, the $\Pi$-Model and Temporal Ensembling~\cite{TemporalEnsembling} showed the effectiveness of updating the network parameters via Exponential Moving Average (EMA). The Mean-Teacher (MT) model~\cite{meanteacher} extended this idea by deploying two networks: the ``student'' and the ``teacher'' networks, both of which have the exact same architecture. In each training iteration, the gradient only back-propagates through the student network, and the parameters in the teacher network are updated by EMA, as shown in Equation~\ref{eq:ema}
\begin{equation}
    \theta_{t}' = (1-\mu)*\theta_{s}'+\mu*\theta_{t}
    \label{eq:ema}
\end{equation}
where $\theta_{t}'$ is the updated teacher parameters, $\theta_t$ is the previous teacher parameters, $\theta_s'$ is the updated student parameters, and $\mu$ is the EMA factor. The divergence between the outputs of the teacher and the student networks is minimized by minimizing the Mean Square Error (MSE) between the predictions of the two networks over training samples, as shown in Equation \ref{eq:mt_consistency}
\begin{equation}
    L_{MT} = \frac{1}{N}\sum_i^N MSE(f_{teacher}(x_i), f_{student}(x_i)).
    \label{eq:mt_consistency}
\end{equation}

\textbf{Label Propagation}: Label propagation is a popular pseudo labeling technique, in which labels propagate from labeled samples to unlabeled samples in their proximity. In~\cite{lowshot}, label propagation was studied in the context of few shot learning. In~\cite{LabelProp2019, walking}, label propagation was applied on SSL. Label propagation with diffusion~\cite{LabelProp2019} deploys a nearest-neighbor graph, which is represented using an affinity matrix. At each iteration, $K$ unlabeled samples are selected based on their proximity to other samples and are assigned pseudo labels with confidence weights, in a process called \textit{diffusion prediction}. We refer the reader to~\cite{LabelProp2019} for more details.

\textbf{FixMatch}: FixMatch \cite{fixmatch} combines the consistency loss and pseudo labeling in one training strategy. As the training process progresses, the entropy of the prediction for unlabeled data decreases. Once the prediction probability of a given sample for a certain class exceeds a threshold $\tau$, FixMatch uses that most probable class to pseudo-label the sample. Consistency loss is applied by minimizing the cross entropy between the assigned pseudo label and the prediction of a \textit{hard-augmented} variant of the sample.
The concept of \textit{hard augmentation} is a critical component of FixMatch. Two different augmentation techniques are employed~\cite{cubuk2018autoaugment}: CTAugment~\cite{berthelot2019remixmatch}, which learns the best augmentation from data, and RandAugment~\cite{cubuk2019randaugment}, in which the augmentation is randomly selected from a pool. FixMatch applies EMA to update the network parameters. Its initial learning rate is small compared to other state of the art methods~\cite{meanteacher, LabelProp2019}. As a result, FixMatch requires $2^{20}$ training iterations to achieve its good performance, which is much larger than other methods.




%% file: sections/04_method.tex
\section{Proposed Semi-supervised Learning Method}
\label{sec:method}
In SSL, the training dataset $X$ is divided into two parts: $X_l$ for the labeled data, where $Y_l$ represents its labels, and $X_u$ for the unlabeled data. The task is to learn features from $X = X_u \cup X_l $ leveraging $Y_l$. The key part of \mbox{MUSCLE} is involving USL from the very beginning  so that we can  extract meaningful features from $X_u$.
In this section, we will first introduce the concept of \mbox{MUSCLE}, and then discuss its properties, functionality, and key aspects. 

\subsection{The Objective of MUSCLE}
\mbox{MUSCLE} literally comes from the idea of training USL with semi-supervised or supervised Learning using Mutual Information (MI) \cite{mutualInfomation} maximization, which has been proved to be useful in both \textit{representation learning} and USL tasks \cite{IMSAT, DeepInfoMap, iic}. Similar to \cite{iic}, \mbox{MUSCLE} applies the Mutual Information Loss (MIL) to the network's likelihood prediction as shown in Equation \ref{eq:MIloss}
\begin{equation}
    l_u = I(f_\theta(x_\alpha), f_\theta(x_\beta))\label{eq:MIloss}
\end{equation}
where $f_\theta$ is the classification function, and $x_\alpha$ and $x_\beta$ are the augmented data of $x$ through transformation functions $A(x)$ and $B(x)$, respectively. The MI is calculated as shown in Equation \ref{eq:MI}~\cite{mutualInfomation, iic}
\begin{equation}
    I(z,z') = I(P) = \sum_{c=1}^{C}\sum_{c'=1}^{C}P_{cc'}\ln{\frac{P_{cc'}}{P_{c}P_{c'}}}
\label{eq:MI}
\end{equation}
where
\begin{equation}
    P=\frac{Q+Q^T}{2}, \enspace Q = \frac{1}{n}\sum_{i=1}^{N}f_{\theta}(x_i) \times f_{\theta}(x_i')^T
\end{equation}
where $C$ is the number of classes, $P$ is a $C \times C$ symmetric matrix, $P_{cc'}$ is the value at the $c^{th}$ row and $c'^{th}$ column of $P$, $P_c$ and $P_{c'}$ are the summations over the $c^{th}$ row and the $c'^{th}$ column, respectively. The total loss function becomes
\begin{equation} \label{muscle_loss}
    L_{MUSCLE} = l_s - \alpha l_u
\end{equation}
where $l_s$ can be any supervised loss from either real or pseudo labels, $l_u$ is the MI between different outputs with the same base sample $x$, and  $\alpha$ is the factor of the MIL. From Equation \ref{muscle_loss}, we can see that the loss is minimized when the MIL term $\alpha l_u$ is maximized. One of MUSCLE's advantages is that it could be combined with other existing SSL models or losses to mitigate their weakness instead of merely replacing them. For example, since pseudo labeling methods provide pseudo labels for supervised classification, they can be used with MUSCLE under the $l_s$ loss term. Any consistency loss $l_c$ can be also added to Equation~\ref{muscle_loss} as:
\begin{equation} \label{total_loss}
    L = L_{MUSCLE} + \beta \l_c = l_s - \alpha l_u + \beta \l_c
\end{equation}
In this work, we use Label Propagation (LP)~\cite{LabelProp2019}, the Mean Teacher model (MT)~\cite{meanteacher}, and FixMatch~\cite{fixmatch} as  base methods to highlight the advantage of combining other SSL techniques with MUSCLE.

\subsection{Properties Of MUSCLE}
\textbf{Avoiding the Trivial Solutions}: The reason that consistency loss does not have the capability of learning meaningful features from $X_u$ without the prior knowledge generated by $X_l$ and $Y_l$ is that the network can simply output the same prediction for all classes of the input data, e.g. $[1,0,\dots,0]$. In such a case, the consistency loss is zero but the solution is obviously meaningless. However, when maximizing the MI, this trivial solution is avoided. 
MI (Equation ~\ref{eq:MI}) can be expended to:
\begin{equation} \label{mutual_information}
    I(z,z') = H(z) - H(z|z')
\end{equation}
where $H(z)$ is the entropy of $z$, or in other words, how much information $z$ contains, and $H(z|z')$ is the conditional entropy of $z$ given $z'$. Therefore, when maximizing the MI, the trivial solution is avoided because $H(z)$ is maximized when the average prediction for each class across the batch is the same. Thus, producing a fixed prediction for all samples should not maximize $H(z)$ with one exception: such fixed prediction has the same value for all classes, e.g. $[0.1,0.1,\dots,0.1]$, which is avoided using $H(z|z')$. The necessary condition for minimizing $H(z|z')$ is when the samples' likelihood in $z$ reach one-hot. Exceptions such as  $[0.1,0.1,\dots,0.1]$ increase  $H(z|z')$ and should not exist in the optimal solution. Furthermore, the $l_s$ term in the \mbox{MUSCLE} also acts as a stabilizer to MIL because $l_s$ directly leads labeled data to meaningful predictions, and indirectly affects the unlabeled data since images within the same class are correlated with each other. The usage of maximizing the MI is also discussed in \cite{iic}.


\textbf{Incorporating MUSCLE into Other Approaches:} By taking a closer look at the MI, we can see that maximizing the MI behaves similar to other SSL approaches with the ability of directly classifying unlabeled data $X_u$. It is easy to see that Maximizing the MI is an indirect method of doing pseudo-labeling since the prediction likelihood $z$ will converge to one-hot for minimizing the second term in Equation~\ref{mutual_information}, $H(z|z')$. According to Equation~\ref{consistency_loss}, consistency loss attempts to minimize the differences between two different predictions based on the same input data $x_u$ without knowing its label $y$. For example, using the  Euclidean Distance 
\begin{equation}
    d(z,z') = \sqrt{\sum(z_i-z'_i)^2} \enspace ,
\end{equation}
the distance between $z$ and $z'$ reaches its minimum value of zero if and only if $z=z'$. Using MI, both $z$ and $z'$ should converge to a one-hot vector. Furthermore, since $z$ and $z'$ are the predictions based on the same base sample $x$ using similar or same network architectures and parameters, $z$ and $z'$ generally yield the same one-hot prediction due to the invariance behavior typical of DNNs. Therefore, in a sense maximizing the MI is equivalent to minimizing the consistency loss. The reason that MUSCLE can be combined with different SSL approaches is that they share common optimization goals. Thus, they can help each other for achieving those goals instead of competing with each other for different objectives. 

\subsection{Batch Composition} \label{batch_composition}
The batch for each training iteration can be expressed as: $[x_{u1},\dots,x_{uI}, x_{l1}, \dots, x_{lJ}]$, where $x_{ui}$  and $x_{lj}$ indicate unlabeled and labeled data samples, respectively. Each batch contains $I$  unlabeled data and $J$  labeled data with the ratio of $r = \frac{I}{J}$. The ratio $r$ is a critical parameter for \mbox{MUSCLE}, because MI attempts to predict each sample as one-hot while maintaining the predictions as a uniform distribution over the batch. Since we are randomly drawing data from the dataset, if the dataset itself is nearly balanced, then the selected data for each batch should also follow a uniform distribution over the classes. If we include $J$ labeled data $[x_{li},\dots,x_{lJ}]$ in a training batch of size $B$, where the predictions $z_{lj}\forall j$ already converged to correct one-hot vectors based on the supervised learning term, we are revealing $\frac{J}{B}$ of correct answers to the MI term to learn the remaining samples in that batch. Therefore, $r$ represents the balance of the batch's difficulty for MIL. A good $r$ can prevent the batch from being overly "easy" or overly "hard". Section~\ref{sec:eval} includes an ablation study on the selection of $r$.

\begin{table*}[]
\begin{center}
\setlength{\tabcolsep}{1.2em} 
{\renewcommand{\arraystretch}{1.0}
	\resizebox{\linewidth}{!}{\begin{tabular}{ccccc}
\hlineB{3}
Dataset                                & \multicolumn{4}{c}{CIFAR10}                                         \\ \cline{2-5} 
Num of Labeled Images                  & 1000 (\red{2\%})             & 500 (\red{1\%})               & 250 (\red{0.5\%})$^\dagger$               & 100 (\red{0.2\%})$^\dagger$         \\ \hline
\multicolumn{1}{c|}{Supervised Learning}     & $59.97\pm0.87$          & $51.08\pm1.02$          & $45.74\pm1.92$          & $32.55\pm2.02$    \\ \hline
\multicolumn{1}{c|}{Label Propagation~\cite{LabelProp2019}} & $77.98\pm0.88$          & $67.60\pm1.80$           & $59.55\pm2.58$          & $35.78\pm3.98$   \\
\multicolumn{1}{c|}{Mean-Teacher~\cite{meanteacher}}      & $80.90\pm0.51$           & $72.55\pm2.64$          & $61.26\pm1.96$          & $39.56\pm1.73$    \\
\multicolumn{1}{c|}{LP+MT~\cite{LabelProp2019}}             & $83.07\pm0.70$          & $75.98\pm2.44$          & $64.19\pm1.95$          & $41.62\pm3.50$    \\ \hline
\multicolumn{1}{c|}{MUSCLE}              & $85.46\pm0.85$          & $79.01\pm0.99$          & $70.37\pm1.98$          & $52.79\pm4.81$    \\
\multicolumn{1}{c|}{MUSCLE+MT}           & $86.42\pm0.27$          & $82.02\pm0.22$           & $75.86\pm3.2$          & $\mathbf{59.57\pm4.53}$    \\
\multicolumn{1}{c|}{\textbf{MUSCLE+MT+LP}}    & $\mathbf{86.71\pm0.36}$          & $\mathbf{83.36\pm0.43}$           & $\mathbf{76.46\pm3.05}$        & $59.03\pm3.17$ \\ 
\hlineB{3}
\end{tabular}}
}
\caption{Comparison with the SOTA methods on CIFAR-10 with 13-Layer CNN. Average accuracy and standard deviation are reported. The percentage of the labeled data, w.r.t the entire training dataset, is listed following the number of labels.\\
\footnotesize{$^\dagger$Baseline results were generated by us.}
} \label{normal_cifa10}
\end{center}
\end{table*}

\begin{table*}[]
\begin{center}
\setlength{\tabcolsep}{1.0em} 
{\renewcommand{\arraystretch}{1.0}
	\resizebox{\linewidth}{!}{\begin{tabular}{cccccc}
\hlineB{3}
Dataset                                & \multicolumn{5}{c}{CIFAR-100}                                                                \\ \cline{2-6} 
Num of Labeled Images                  & 10000 (\red{20\%})            & 4000 (\red{8\%})             & 2500 (\red{5\%})$^\dagger$             & 500 (\red{1\%})$^\dagger$              & 100 (\red{0.2\%})$^\dagger$               \\ \hline
\multicolumn{1}{c|}{Supervised Learning}     & $59.33\pm0.49$          & $44.57\pm0.11$          & $36.69\pm0.70$          & $16.02\pm0.53$          & $5.74\pm0.64$           \\ \hline
\multicolumn{1}{c|}{Label Propagation~\cite{LabelProp2019}} & $61.57\pm1.88$          & $53.8\pm0.76$           & $48.84\pm0.38$          & $17.49\pm1.01$          & $4.94\pm0.42$           \\
\multicolumn{1}{c|}{Mean-Teacher~\cite{meanteacher}}      & $63.92\pm0.51$          & $54.64\pm0.49$          & $47.28\pm0.82$          & $20.45\pm0.61$          & $6.84\pm0.89$           \\
\multicolumn{1}{c|}{LP+MT~\cite{LabelProp2019}}             & $64.08\pm0.47$          & $56.27\pm0.20$          & $51.14\pm0.44$          & $21.40\pm0.68$           & $5.66\pm0.84$           \\ \hline
\multicolumn{1}{c|}{\textbf{MUSCLE+MT}}           & $\mathbf{66.07\pm0.19}$         & $\mathbf{59.31\pm0.45}$           & $\mathbf{54.53\pm0.35}$     & $\mathbf{29.86\pm0.85}$        & $\mathbf{11.01\pm0.85}$ \\
\multicolumn{1}{c|}{MUSCLE+MT+LP}    & $64.79\pm0.25$          & $57.66\pm0.62$          & $52.48\pm0.46$          & $28.27\pm0.73$          & $10.51\pm0.43$          \\ \hlineB{3}
\end{tabular}}
}
\caption{Comparison with SOTA methods on CIFAR-100 with 13-Layer CNN. Average accuracy and standard deviation are reported. The percentage of the labeled data, w.r.t the entire training dataset, is listed following the number of labels.\\ \footnotesize{$^\dagger$Baseline results were generated by us.}}
\label{normal_cifa100} 
\end{center}
\end{table*}

\subsection{Data Augmentation}
The effectiveness of the data augmentation in SSL has been well studied \cite{miyato2017virtual, UDA, fixmatch}. Often, only one augmentation function is used, which can be light augmentation \cite{meanteacher, LabelProp2019} or hard augmentation \cite{iic}. In~\cite{fixmatch}, it was shown that using both light and hard augmentations into the consistency loss can have a much better result because it creates a larger divergence for the consistency loss to achieve better generalization. We also adopt the concept of two types of augmentation where the easy one is the classical augmentation used in \cite{meanteacher} and the hard one is either the augmentation used in \cite{iic} without sobel processing or the RandAugment \cite{cubuk2019randaugment}.

\begin{table*}[]
\begin{center}
\setlength{\tabcolsep}{0.4em} 
{\renewcommand{\arraystretch}{1.0}
	\resizebox{\linewidth}{!}{\begin{tabular}{ccccccc}
\hlineB{3}
Dataset                                & \multicolumn{6}{c}{Mini-ImageNet}                                                                                                    \\ \cline{2-7} 
                                       & \multicolumn{3}{c|}{Top 1 Accuracy}                                         & \multicolumn{3}{c}{Top 5 Accuracy}                     \\
Num of Labeled Images                  & 10000 (\red{20\%})             & 4000 (\red{8\%})            & \multicolumn{1}{c|}{2500 (\red{5\%})$^\dagger$}             & 10000 (\red{20\%})            & 4000 (\red{8\%})             & 2500 (\red{5\%})$^\dagger$             \\ \hline
\multicolumn{1}{c|}{Supervised Learning}     & $39.63\pm0.53$          & $25.62\pm0.36$          & \multicolumn{1}{c|}{$19.01\pm0.71$}          & $61.39\pm0.46$          & $44.26\pm0.45$          & $35.48\pm0.73$          \\ \hline
\multicolumn{1}{c|}{Label Propagation~\cite{LabelProp2019}} & $45.47\pm0.47$          & $29.71\pm0.69$          & \multicolumn{1}{c|}{$22.41\pm0.62$}                & $69.46\pm0.31$          & $52.74\pm0.49$          & $43.13\pm1.01$                \\
\multicolumn{1}{c|}{Mean-Teacher~\cite{meanteacher}}      & $44.82\pm0.51$          & $28.23\pm0.23$           & \multicolumn{1}{c|}{$22.34\pm0.52$}          & $69.38\pm0.95$          & $51.94\pm0.42$          & $44.61\pm0.78$          \\
\multicolumn{1}{c|}{LP+MT~\cite{LabelProp2019}}             & $45.92\pm0.37$            & $28.67\pm0.41$                & \multicolumn{1}{c|}{$23.38\pm0.35$}                & $70.99\pm0.56$                & $52.26\pm0.63$                & $45.67\pm0.86$                \\ \hline
\multicolumn{1}{c|}{MUSCLE}              & $45.86\pm0.06$          & $34.90\pm0.47$          & \multicolumn{1}{c|}{$28.95\pm0.45$}          & $71.26\pm0.10$          & $60.81\pm0.21$          & $54.45\pm0.83$          \\
\multicolumn{1}{c|}{\textbf{MUSCLE+MT}}           & $\mathbf{49.47\pm0.30}$     & $\mathbf{38.26\pm0.15}$ & \multicolumn{1}{c|}{$\mathbf{32.56\pm0.12}$} & $\mathbf{72.94\pm0.29}$ & $\mathbf{63.35\pm0.21}$ & $\mathbf{56.58\pm0.29}$ \\
\multicolumn{1}{c|}{MUSCLE+MT+LP}    & $47.30\pm1.12$     & $37.35\pm0.25$        & \multicolumn{1}{c|}{$30.86\pm0.53$}                & $71.06\pm0.64$                & $\mathbf{63.84\pm0.53}$                & $\mathbf{56.79\pm0.48}$ \\ \hlineB{3}
\end{tabular}}
}
\caption{Comparison with the SOTA Methods on Mini-Imagenet with Resnet18. Average accuracy and standard deviation are reported. The percentage of the labeled data, w.r.t the entire training dataset, is listed following the number of labels.\\ \footnotesize{$^\dagger$Baseline results were generated by us.}}
\label{normal_miniimagenet}
\end{center}
\end{table*}

%% file: sections/05_eval.tex
\section{Experimental Evaluation}
\label{sec:eval}
We first present the benchmark datasets used for the SSL evaluation, followed by the network structure and hyper-parameters settings. Then, we present a comparison between \mbox{MUSCLE} and the state-of-the-art methods. We also introduce a set of ablation studies on several key factors of MUSCLE. Finally, we include experiments to demonstrate an explanation for \mbox{MUSCLE}'s main strengths.

\subsection{Benchmarks Dataset}
We conducted experiments on CIFAR-10~\cite{cifar}, CIFAR-100~\cite{cifar}, and Mini-Imagenet~\cite{miniimagenet}. We put a special emphasis on the performance when the amount of  labeled data is significantly reduced to showcase \mbox{MUSCLE}'s clear advantage in data-starved scenarios. For example, for CIFAR-10, while experiments in~\cite{meanteacher} used a minimum of 1000 labeled data samples, and in~\cite{LabelProp2019} used a minimum of 500 labeled data samples, in our experiments we included evaluations on only 250 and 100 labeled data samples.


\textbf{CIFAR-10 and CIFAR-100}: Both CIFAR10 and CIFAR100 contain 60K of $32 \times 32$ RGB images, from 10 and 100 classes, respectively. In both datasets, all classes have the same number of samples, $\frac{1}{6}$ of which is dedicated for testing and the rest is for training. For CIFAR-10, we randomly selected 100, 50, 25, and 10 samples from each class to form the labeled dataset. For CIFAR-100, we randomly selected 100, 40, 25, 5, and 1 images from each class to form the labeled dataset. We use the rest of the training data as unlabeled samples.

\textbf{Mini-Imagenet}: Mini-Imagenet~\cite{miniimagenet} 
is a subset of ImageNet~\cite{imagenet_cvpr09} that contains 60K $84 \times 84$ 3-channel images from 100 classes. However, different from normal classification datasets, it is split into 64-16-20 classes, where 64 classes are for training, 16 classes for validating, and 20 classes for testing. To evaluate SSL on this dataset, we followed the approach used in \cite{LabelProp2019}. For each class, we randomly assigned 500 images to training and 100 images to testing. In total, we used 50K images for training and 10K images for testing. Then, we randomly selected 100, 40, and 25 images from each class to form the labeled samples, and use the rest of the training data as unlabeled samples. 


\begin{table*}[]
\begin{center}
\setlength{\tabcolsep}{0.1em} 
{\renewcommand{\arraystretch}{1.0}
	\resizebox{\linewidth}{!}{\begin{tabular}{cccccccc}
\hlineB{3}
Dataset                            & \multicolumn{3}{c}{CIFAR-10}                                                & \multicolumn{4}{c}{CIFAR-100}                                            \\ \cline{2-8} 
Num of Labeled Images              & 500 (\red{1\%})              & 250 (\red{0.5\%})               & \multicolumn{1}{c|}{100 (\red{0.2\%}) }              & 10000 (\red{20\%})             & 4000 (\red{8\%})             & 2500 (\red{5\%})              & 500 (\red{1\%})               \\ \hline
\multicolumn{8}{c}{Hyper-Parameter Setting 1}                                                                                                                                                \\ \hline
\multicolumn{1}{c|}{Fixmatch~\cite{fixmatch}}      & $84.12\pm0.55$          & $81.96\pm0.75$          & \multicolumn{1}{c|}{$73.98\pm1.46$}          & $53.35\pm0.72$          & $45.37\pm0.56$          & $41.02\pm0.31$          & $21.02\pm0.84$          \\
\multicolumn{1}{c|}{\textbf{MUSCLE+FixMatch}} & $\mathbf{84.59\pm0.61}$ & $\mathbf{82.63\pm0.78}$ & \multicolumn{1}{c|}{$\mathbf{76.06\pm1.81}$} & $\mathbf{54.97\pm0.24}$ & $\mathbf{47.71\pm 0.71}$ & $\mathbf{43.54\pm0.33}$ & $\mathbf{23.73\pm1.67}$ \\ \hline
\multicolumn{8}{c}{Hyper-Parameter Setting 2}                                                                                                                                                \\ \hline
\multicolumn{1}{c|}{Fixmatch~\cite{fixmatch}}      & $90.53\pm0.62$          & $89.51\pm0.71$          & \multicolumn{1}{c|}{$81.62\pm1.12$}            & $68.28\pm0.19$          & $62.28\pm0.13$          & $58.19\pm0.31$          & $32.52\pm0.65$          \\
\multicolumn{1}{c|}{\textbf{MUSCLE+FixMatch}} & $\mathbf{90.91\pm0.37}$ & $\mathbf{90.25\pm0.39}$ & \multicolumn{1}{c|}{$\mathbf{83.51\pm1.77}$} & $\mathbf{68.51\pm0.23}$ & $\mathbf{62.66\pm0.18}$ & $\mathbf{58.53\pm0.49}$ & $\mathbf{33.94\pm0.72}$ \\ \hlineB{3}
\end{tabular}}
}
\caption{Comparison with the FixMatch model on CIFAR10 and CIFAR100 datasets with 13-Layer CNN network. Two hyper-parameter settings (Section~\ref{training}) are used. Average accuracy and standard deviation are reported. The percentage of the labeled data, w.r.t the entire training dataset, is listed following the number of labels. All results are generated by us.}
\label{fixmatch_cifar}
\end{center}
\end{table*}

\begin{table*}[]
\begin{center}
\setlength{\tabcolsep}{0.5em} 
{\renewcommand{\arraystretch}{1.0}
	\resizebox{\linewidth}{!}{\begin{tabular}{ccccccc}
\hlineB{3}
Dataset                            & \multicolumn{6}{c}{Mini-ImageNet}                                                                      \\ \cline{2-7} 
                                   & \multicolumn{3}{c|}{Top 1 Accuracy}                          & \multicolumn{3}{c}{Top 5 Accuracy}      \\
Num of Labeled Images              & 10000 (\red{20\%})        & 4000 (\red{8\%})         & \multicolumn{1}{c|}{2500 (\red{5\%}) }        & 10000 (\red{20\%})        & 4000 (\red{8\%})         & 2500 (\red{5\%})         \\ \hline
\multicolumn{7}{c}{Hyper-Parameters Setting 1}                                                                                              \\ \hline
\multicolumn{1}{c|}{FixMatch~\cite{fixmatch}}      & $26.71\pm0.70$ & $18.57\pm0.32$ & \multicolumn{1}{c|}{$14.12\pm0.27$} & $51.14\pm1.80$ & $39.38\pm0.29$ & $32.52\pm1.29$ \\
\multicolumn{1}{c|}{\textbf{MUSCLE+FixMatch}} & $\mathbf{27.33\pm0.19}$ & $\mathbf{18.72\pm0.28}$ & \multicolumn{1}{c|}{$\mathbf{14.58\pm0.29}$} & $\mathbf{52.43\pm0.31}$ & $\mathbf{39.83\pm0.65}$ & $\mathbf{33.16\pm0.51}$ \\ \hline
\multicolumn{7}{c}{Hyper-Parameters Setting 2}                                                                                              \\ \hline
\multicolumn{1}{c|}{FixMatch~\cite{fixmatch}}      & $36.05\pm0.84$ & $25.82\pm0.99$ & \multicolumn{1}{c|}{$19.11\pm0.94$} & $59.26\pm1.03$ & $48.61\pm1.82$ & $39.58\pm1.47$ \\
\multicolumn{1}{c|}{\textbf{MUSCLE+FixMatch}} & $\mathbf{37.71\pm0.48}$ & $\mathbf{29.29\pm0.12}$ & \multicolumn{1}{c|}{$\mathbf{24.73\pm0.68}$} & $\mathbf{62.74\pm0.22}$ & $\mathbf{53.60\pm0.33}$  & $\mathbf{48.23\pm1.05}$ \\ \hlineB{3}
\end{tabular}}
}
\caption{Comparison with the FixMatch model on the Mini-ImangeNet dataset with Resnet18 using two hyper-parameter settings (Section~\ref{training}). The top 1 and top 5 average accuracy and standard deviation are reported. The percentage of the labeled data w.r.t the entire training dataset is listed following the number of labels. All results are generated by us.}
\label{fixmatch_miniimagenet}
\end{center}
\end{table*}

\subsection{Training} \label{training}
We implemented our method in PyTorch~\cite{pytorch} and used the public implementations of LP~\cite{LabelProp2019} and the MT~\cite{meanteacher}. SGD~\cite{sgd} was used to optimize all the models. We also implemented the loss function of FixMatch \cite{fixmatch} for combining \mbox{MUSCLE} with FixMatch.
%

For CIFAR-10 and CIFAR-100, we used the 13-Layer CNN network that was used in~\cite{meanteacher, LabelProp2019}. For Mini-Imagenet, we trained a Resnet18 network for the feature extractor. In a mini-batch, similar to \cite{iic}, we performed hard augmentation on each image three times such that for each original image, a single weakly augmented version can be paired with three hardly augmented versions. This can increase the generality and improve the training stability.



When we compare \mbox{MUSCLE} with MT~\cite{meanteacher} and LP~\cite{LabelProp2019}, we used the hyper-parameters in these methods. The network was trained over 180 epochs and the initial learning rate for MUSCLE was 0.05 for all datasets.  A Cosine Learning Rate decay~\cite{coslrdecay} was used to adjust the learning rate where the learning rate reaches 0 at the 210$^{th}$ epoch. In each training batch, there are 128 images in total, including 64 labeled images. The ratio $r$ (Section\ref{batch_composition}) equals to 1. We followed the baselines' batch compositions and learning rate when \mbox{MUSCLE} is not involved.  

Upon comparing MUSCLE with FixMatch~\cite{fixmatch}, we noticed that FixMatch was trained on a TPU for $2^{20}$ iterations with a total batch size of 512 images, which is far beyond the computing resources available to us. For a fair comparison and demonstrating the potential that FixMatch can benefit from combining with MUSCLE, we evaluated FixMatch and \mbox{MUSCLE} with the same number of training iterations. We used two sets of training parameters: (1) the hyper-parameter settings in the MT and LP models, which were listed above. (2) the hyper-parameter settings in FixMatch with our batch composition and reduced number of iterations, such that the model is trained for 300 epochs with initial learning rate of 0.03. The learning rate is adjusted over a $\frac{7}{16}$ cycle of cosine learning rate decay. We separately listed the comparison with FixMatch to avoid confusion.
 
 \subsection{Comparison with the State of the Art}
 Tables~\ref{normal_cifa10}, \ref{normal_cifa100}, \ref{normal_miniimagenet}, \ref{fixmatch_cifar}, and~\ref{fixmatch_miniimagenet} compare the testing accuracy with supervised learning and the baseline methods~\cite{LabelProp2019, meanteacher} with and without the use of \mbox{MUSCLE}. With \mbox{MUSCLE}, all baseline models consistently achieve better performance on all three datasets and all experimental setups. For comparing with FixMatch, the performance increases on both hyper-parameter settings for all datasets and all experimental setups. It is important to also note that  the accuracy boost achieved with MUSCLE increases as the number of the labeled images decreases. This matches our expectations and show the advantage of \mbox{MUSCLE} in label-starved data scenarios. In Table~\ref{normal_cifa100} and Table~\ref{normal_miniimagenet}, the performance of MT+\mbox{MUSCLE} is better than \mbox{MUSCLE}+MT+LP. The reason for this is that LP follows a two-stage training. The first stage trains the model without using LP for acquiring necessary preliminary knowledge about the task. Then, in the second stage, that knowledge is used to assign pseudo labels. Due to LP's properties, one hypothesis we have is that the second stage needs to start with either a highly accurate model, or a well-calibrated model~\cite{calibrate_model}. However, in the case of a large number of classes (as in CIFAR-100 and Mini-ImageNet), the base model trained by MT+\mbox{MUSCLE} might not sufficiently satisfy either these requirements. Thus, adding another training stage with LP could be counter productive.
 
 
 \subsection{Ablation Study}
\begin{figure*}[t]
\begin{center}
\subfloat[]{
   \includegraphics[width=0.25\linewidth]{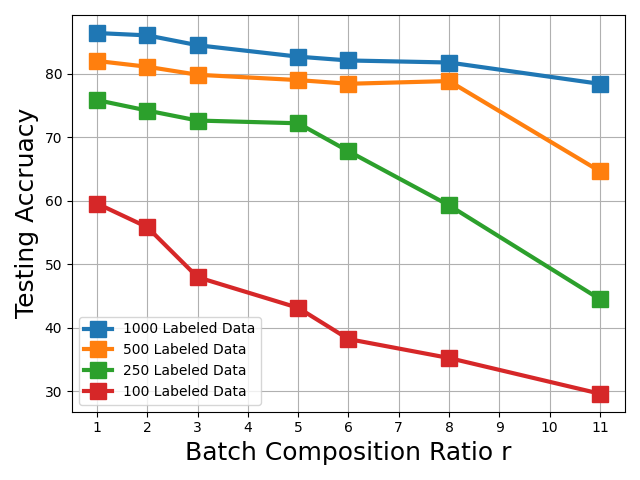}
 }
 \subfloat[]{
   \includegraphics[width=0.25\linewidth]{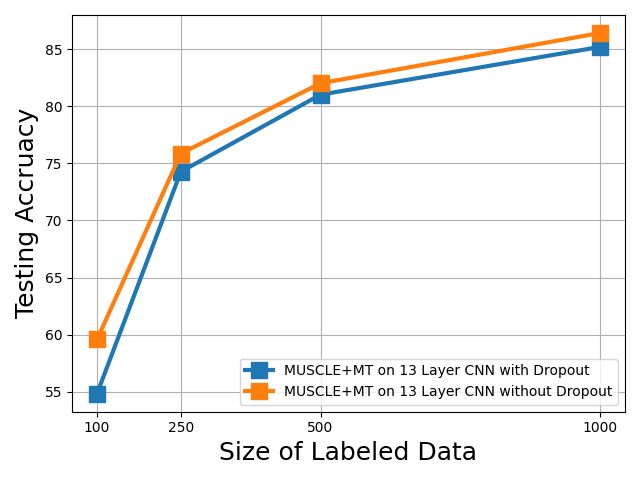}
 }
  \subfloat[]{
   \includegraphics[width=0.25\linewidth]{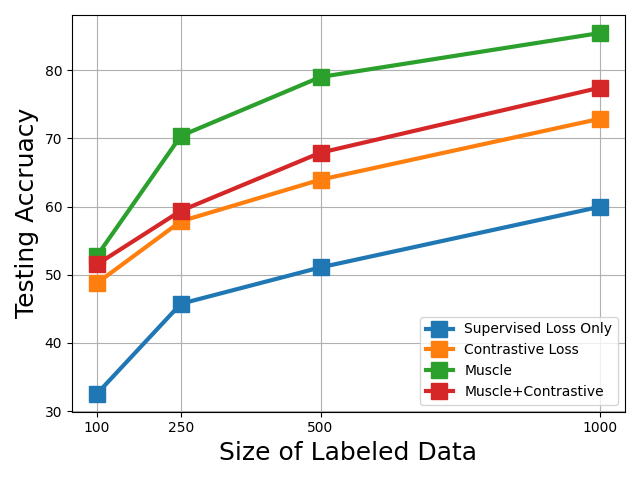}
 }
  \subfloat[]{
   \includegraphics[width=0.25\linewidth]{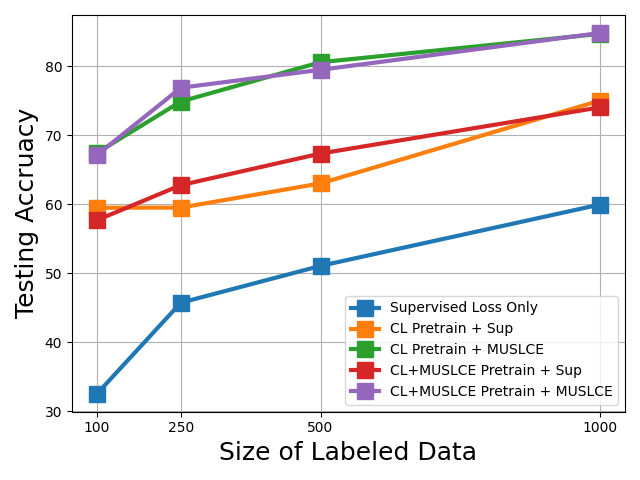}
 }
\end{center}
   \caption{Ablation studies on CIFAR-10: (a) Batch composition ratio $r$ (Section~\ref{batch_composition}) vs.~testing accuracy on MUSCLE+MT, (b) Impact of the dropout layer in the 13-layer CNN network, (c) Comparison on unsupervised learning losses when learning from scratch, (d) Comparison on unsupervised learning losses when performing pre-training followed by fine-tuning.} \label{combined_ablation}
\label{fig:long}
\label{fig:onecol}
\end{figure*}

 \textbf{Impact of $r$ and Dropout Layer: }  We studied the impact of the ratio $r$, explained in Section~\ref{batch_composition}. Figure~\ref{combined_ablation} (a) shows the testing accuracy with different values of $r$ on CIFAR-10. The performance is relatively flat for $1 < r < 2$. When $r$ increases beyond 2, where unlabeled data amount is much larger than the labeled data amount, the accuracy decays significantly. For the Dropout Layer in the 13-layer CNN, MT~\cite{meanteacher} has provided a detailed ablation study for showing its importance to MT model. However, with MUSCLE included, as shown in Figure~\ref{combined_ablation} (b), removing the Dropout layer can provide positive effect. Since most of the commonly used networks (e.g. ResNet or VGG) do not natively contain Dropout layers in the feature extractor, our method can be used with those networks without changing the architecture. 
 

\textbf{Comparison with Contrastive Loss}: To show the benefit of the MIL in MUSCLE compared to simple Contrastive Loss (CL), which is commonly used in self-supervised learning, we provide two extra sets of ablation studies. First, we trained models from scratch on CIFAR10 by either replacing MIL with CL or combining MIL with CL. Second, we pre-trained models on CIFAR10 in an unsupervised learning manner by either only using CL or combining CL with MIL. Then, we fine-tuned them using either supervised learning or MUSCLE.  Results are shown in Figure~\ref{combined_ablation} (c) and (d), respectively. When training models from scratch, merely using CL performs better than the supervised learning baseline, but worse than either CL+MUSCLE or MUSCLE alone. This outcome is easy to understand as the CL considers each individual image as a standalone class. Even if two samples belong to the same class, the loss still attempts to push them away from each other. For the pre-training+fine-tuning experiment, the outcome shows that although adding MUSCLE to the pre-training stage  hardly provides any benefit, adding MUSCLE to the fine-tuning stage clearly boosts the performance.
 
\begin{figure}[t]
\begin{center}
\subfloat[Accuracy with 2000 Label Images]{
   \includegraphics[width=0.5\linewidth]{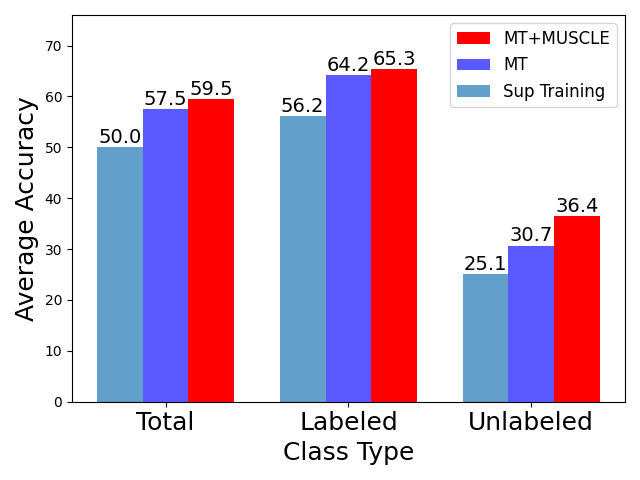}
 }
 \subfloat[Accuracy with 4000 Label Images]{
   \includegraphics[width=0.5\linewidth]{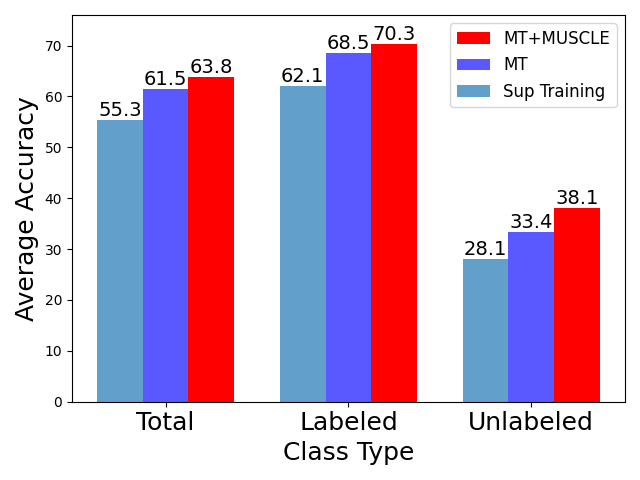}
 }
  \vskip\baselineskip
  \subfloat[Entropy with 2000 Label Images]{
   \includegraphics[width=0.5\linewidth]{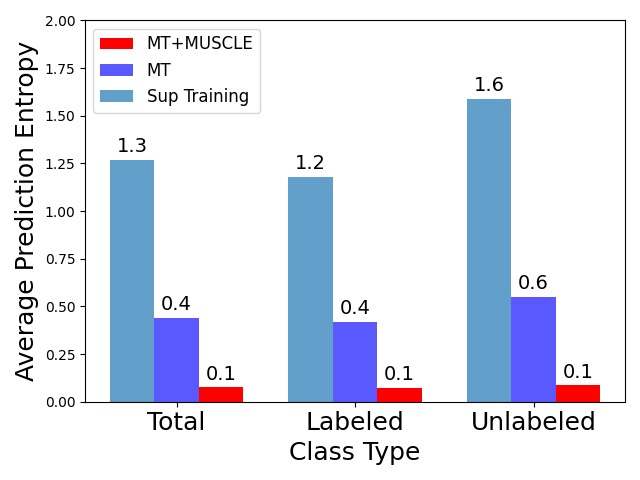}
 }
  \subfloat[Entropy with 4000 Label Images]{
   \includegraphics[width=0.5\linewidth]{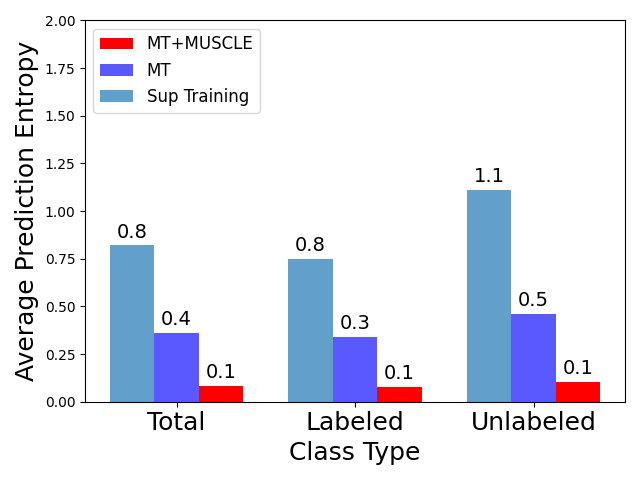}
 }
\end{center}
   \caption{Average accuracy and prediction entropy by different class types on CIFAR-20: (a)-(b) testing accuracy, (c)-(d) average prediction entropy.} \label{cifar20_figure}
\label{fig:long}
\label{fig:onecol}
\end{figure}

 \textbf{Sequestered Classes:} The key point we claim for MUSCLE to work is that, compared with other SSL methods, MUSCLE can directly learn meaningful representations from unlabeled data due to involving USL early-on in the training process. Therefore, MUSCLE should have an advantage when the labeled data does not have enough samples for representing a specific class. For verifying this claim, we experimented on CIFAR-20, a hierarchical dataset based on the CIFAR-100. CIFAR-20 groups the 100 classes from CIFAR-100 into 20 super-classes, with each super-class having five sub-classes. In the same super-class, although the images from different sub-classes share some similarity, it is very hard to infer the super-class of a image based on another image from a different sub-class. For example, dolphins and otters both belong to aquatic mammals, but it is hard to classify a dolphin to aquatic mammal by only knowing otters are aquatic mammals. In this case, we introduce a new experimental setup. We randomly select a sub-class for each super-class and completely remove all label information for that sub-class and call them \textit{unlabeled class}. Then, we randomly select $k$ images from the rest of the classes to form the labeled dataset and call them \textit{labeled classes}. In other words, an \textit{unlabeled class} will not contribute to the labeled data but they still contribute to the unlabeled data. By following this setup, we believe that the selected labeled images cannot fully represent their super-classes. In Figure~\ref{cifar20_figure}, we can see that, compared with supervised learning and MT, \mbox{MUSCLE} delivers a performance boost in all three types of classes, but the majority of the performance improvement is in the unlabeled class. Furthermore, the entropy of the predictions on testing data shows that for both supervised learning method and MT model, the predictions can be affected by the class type and the amount of labeled data, whereas \mbox{MUSCLE} has a very constant prediction entropy across all class types and label amounts.

%% file: sections/06_conclusion.tex
\section{Conclusion}
We presented Mutual-information-based Unsupervised and Supervised Concurrent LEarning (\mbox{MUSCLE}), which is a powerful framework for semi-supervised learning that combines the merits of leading SSL and USL techniques. In contrast to prior attempts, MUSCLE involves USL in the training process from the first iteration. MUSCLE achieved consistent improvement over the state of the art over three standard datasets, across all experimental setups. The performance boost gained by MUSCLE is maximum when the amount of training data is lowest, e.g. one sample per class for CIFAR-100. MUSCLE's power is further underscored by its extra robustness in the situation when the labeled data is biased. Finally, MUSCLE exhibited significant potential in fine-tuning pre-trained models.